\documentclass[runningheads]{llncs}

\usepackage{eccv}

\usepackage{eccvabbrv}

\usepackage{graphicx}
\usepackage{booktabs}
\usepackage{wrapfig}
\usepackage{stmaryrd}
\usepackage{booktabs}
\usepackage{multirow}
\usepackage{amssymb}
\usepackage{subcaption}

\usepackage[accsupp]{axessibility}  

\usepackage{hyperref}

\usepackage{orcidlink}

\usepackage{pifont}

\begin{document}

\title{SHReg: Strictly Rotation-Equivariant Point Cloud Registration via Spherical Harmonics} 
\titlerunning{SHReg}

\author{
Chongjian Wang\inst{1}\orcidlink{0009-0008-7204-3225}
\and
Junjie Gao\inst{2}\orcidlink{0000-0001-7087-0886}\thanks{Corresponding author.}
}

\authorrunning{C.~Wang and J.~Gao}

\institute{
College of Mathematics and Systems Science, Shandong University of Science and Technology, Qingdao, China\\
\email{202311080223@sdust.edu.cn}
\and
School of Artificial Intelligence, Shandong Women's University, Jinan, China\\
\email{gjjsdnu@163.com}
}

\maketitle

\begin{abstract}
Point cloud registration critically depends on local features that are both distinctive and robust to arbitrary 3D rotations. Existing learning-based methods typically approximate rotation invariance via fragile local reference frames or extensive data augmentation, providing only empirical invariance and often degrading under unseen rotational transformations.
In this paper, we propose SHReg, a strictly rotation-equivariant point cloud registration framework grounded in the representation theory of $SO(3)$. By representing local geometric features as irreducible representations of $SO(3)$, SHReg guarantees exact equivariance under arbitrary rotations without relying on local reference frames. Built upon a spherical-harmonics-based equivariant backbone, SHReg jointly learns rotation-invariant descriptors for robust correspondence matching and rotation-equivariant features that preserve fine-grained orientation information.
The preserved equivariant structure enables each correspondence to directly hypothesize a rigid transformation, reducing reliance on large-scale hypothesis sampling in conventional RANSAC-based pipelines and leading to improved robustness under challenging rotational variations.
Extensive experiments on 3DMatch, 3DLoMatch, and KITTI demonstrate that SHReg consistently outperforms state-of-the-art methods in registration accuracy, particularly under large rotational perturbations.

\end{abstract}

\section{Introduction}
\label{sec:intro}
Point cloud registration, which estimates a rigid transformation to align two partially overlapped point clouds, is a fundamental problem in 3D computer vision. It has been widely applied in autonomous driving\cite{li2020deep, Yue2018ALP}, robot localization\cite{Ullah2024MobileRL}, and 3D reconstruction\cite{Liu2021VoteHMROV,Jaramillo2024CulturalH3,Xu2024GRMLG}. Among various paradigms, the feature-based registration framework remains one of the most actively studied directions, typically consisting of two key steps: local descriptor extraction\cite{Rusu2009FastPF, Salti2014SHOTUS, Deng2018PPFNetGC, Deng2018PPFFoldNetUL, Choy2019FullyCG,Wang2021YouOH, Wang2023RoRegPP, Gao2023OAAFormerRA, Yu2023RotationInvariantTF, Yao2024PARENetPR} and robust transformation estimation\cite{Li2020GESACRG,Hu2014Super4F,Fischler1981RandomSC,Bai2021PointDSCRP,Jiang2023RobustOR}. A long-standing contradiction in this task lies in the fact that point clouds can exhibit large pose variations, while the extracted descriptors are expected to be invariant to such changes, especially global rotations. Therefore, learning discriminative yet rotation-invariant features is crucial for reliable correspondence establishment and accurate pose recovery.

To obtain rotation-invariant descriptors, existing methods can be roughly categorized into patch-wise\cite{Yew20183DFeatNetWS, Gojcic2018ThePM, Ao2020SpinNetLA, Deng2018PPFNetGC, Deng2018PPFFoldNetUL} and scene-wise feature extractors. Patch-wise approaches learn descriptors from local neighborhoods and usually rely on explicit preprocessing to eliminate orientation ambiguity. Two representative strategies are: (1) using handcrafted rotation-invariant statistics such as distances, angles, or point-pair features\cite{Deng2018PPFNetGC, Deng2018PPFFoldNetUL}; and (2) estimating a local reference frame\cite{Gojcic2018ThePM, Ao2020SpinNetLA} to normalize the local patch before feature extraction. Although these designs can reduce sensitivity to rotations, the former often suffers from information loss due to aggressive invariant summarization, while the latter is vulnerable to imperfect reference frame estimation under partial overlap. In addition, patch-wise methods are typically inefficient because they process a large number of local patches independently, without effectively sharing intermediate computations across the whole point cloud.

Scene-wise extractors\cite{Thomas2019KPConvFA, Choy2019FullyCG, Wang2021YouOH,Wang2023RoRegPP,Yu2023RotationInvariantTF, Yao2024PARENetPR}, on the other hand, compute dense descriptors for all points in a unified forward pass and have achieved strong performance in recent learning-based pipelines. However, most scene-wise methods are built upon rotation-sensitive backbones (e.g., KPConv\cite{Thomas2019KPConvFA}, FCGF\cite{Choy2019FullyCG}) and enforce rotation invariance mainly through extensive rotation augmentation during training\cite{Yu2023RotationInvariantTF}. This practice introduces several drawbacks. First, the continuous rotation space is difficult to cover sufficiently, leading to fragility when encountering unseen orientations. Second, learning rotation invariance\cite{Wang2021YouOH,Wang2023RoRegPP} implicitly often increases model capacity requirements, since the network must “memorize” invariance patterns from data. Third, forcing a rotation-sensitive architecture to become invariant can compromise its ability to focus on learning structural distinctiveness, which limits descriptor discriminability under large pose changes.

\begin{figure}[htbp]
  \centering
  \vspace{-20pt}

  \begin{subfigure}{0.49\linewidth}
    \centering
    \includegraphics[width=\linewidth]{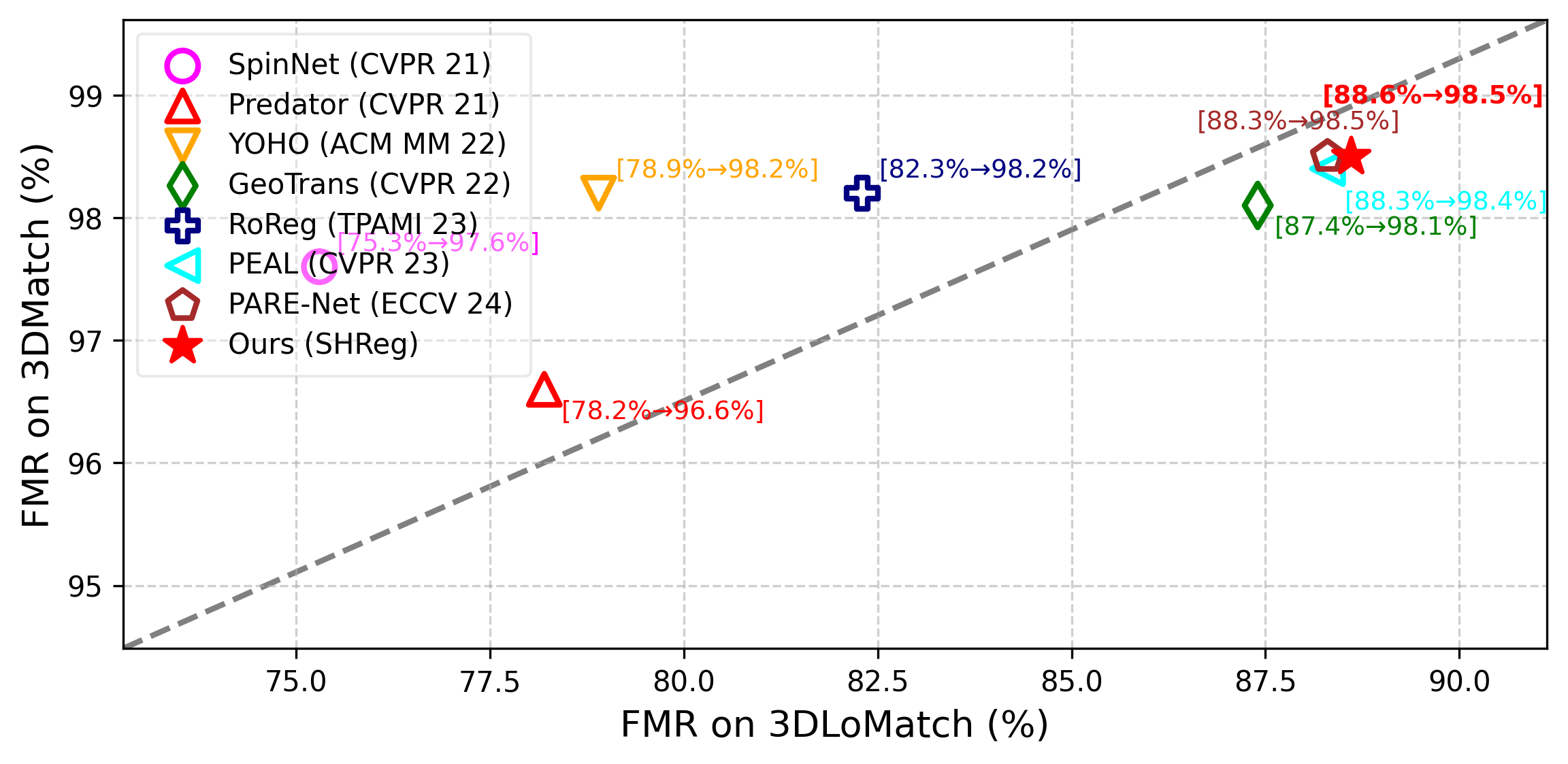}
  \end{subfigure}
  \hfill
  \begin{subfigure}{0.49\linewidth}
    \centering
    \includegraphics[width=\linewidth]{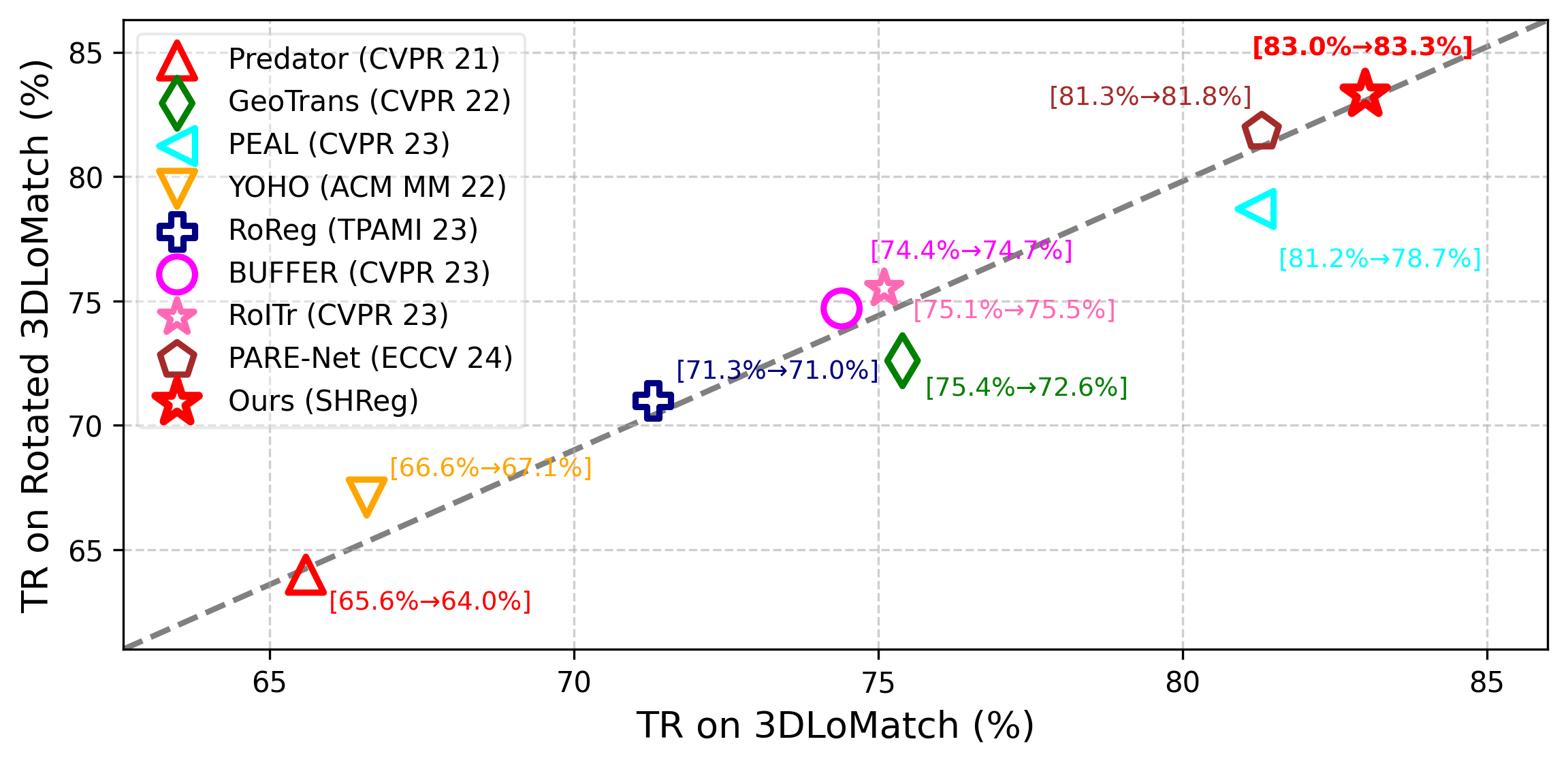}
  \end{subfigure}
  \vspace{-10pt}
\caption{Experimental results on 3DMatch, 3DLoMatch, and Rotated 3DLoMatch.  Our method consistently achieves higher feature matching recall (FMR) and transformation recall (TR), demonstrating strong robustness to rotations.}

  \label{fig:1}
  \vspace{-20pt}
\end{figure}

In this paper, we propose SHReg, a strictly rotation-equivariant point cloud registration framework built upon spherical harmonic representations. The core idea is to explicitly model the $SO(3)$ symmetry of point clouds at the representation level and leverage it consistently in both feature extraction and transformation estimation. Unlike augmentation-driven designs that attempt to approximate rotation invariance from data, SHReg enforces equivariance by construction, enabling the network to produce representations that transform predictably under arbitrary rotations while preserving fine-grained geometric structures. As shown in Fig.~\ref{fig:1}, SHReg consistently achieves higher feature matching recall (FMR), which measures the percentage of scan pairs with sufficiently accurate correspondences, and transformation recall (TR), which evaluates the success rate of pose estimation, across 3DMatch, 3DLoMatch, and their rotated variants.

First, we introduce a spherical harmonic–based strictly rotation-equivariant feature extraction network. Spherical harmonics naturally form a complete and orthogonal basis for functions defined on the sphere and exhibit well-defined transformation behavior under rotations. By embedding local neighborhood geometry into the spherical harmonic domain, SHReg ensures that intermediate representations follow strict equivariant transformation rules. Rotation-invariant descriptors are then derived from these equivariant representations without sacrificing structural distinctiveness. In contrast to rotation-sensitive backbones trained with extensive augmentation, our design builds rotation consistency directly into the architecture, allowing the network to focus on learning intrinsic geometric characteristics rather than compensating for pose variations.
\begin{wrapfigure}{r}{0.45\linewidth}
  \centering
  \vspace{-5pt}
  \includegraphics[width=\linewidth]{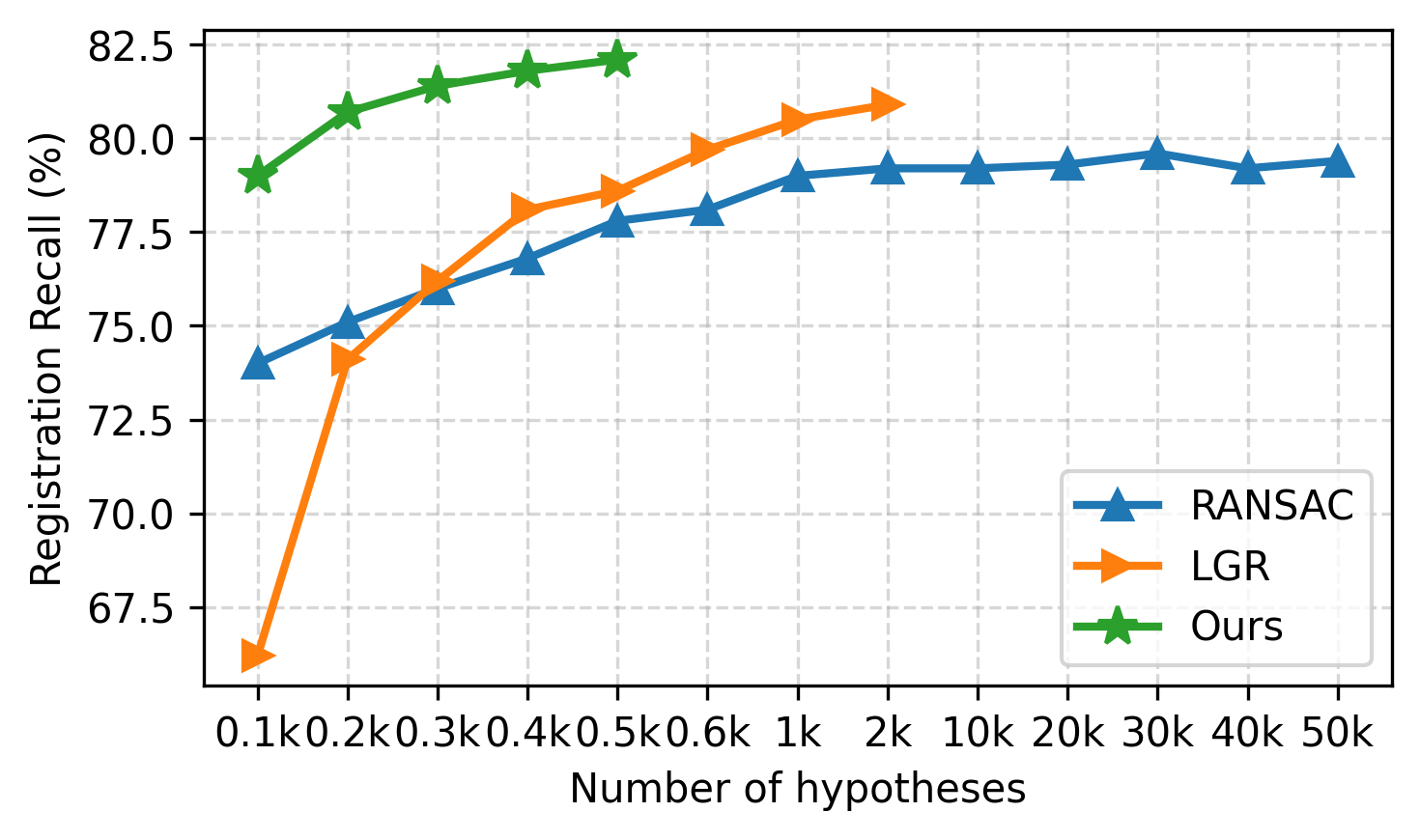}
  \vspace{-15pt}
  \caption{Comparison of our hypothesis proposer, RANSAC and LGR on 3DLoMatch.}
  \label{fig:2}
  \vspace{-20pt}
\end{wrapfigure}

Second, we design a correspondence-driven rigid transformation estimation strategy that directly exploits the consistency of strictly rotation-equivariant features. Once reliable correspondences are established through invariant matching, the equivariant representations provide structured geometric cues for accurate rotation recovery, followed by stable translation estimation. Unlike conventional estimators such as RANSAC\cite{Fischler1981RandomSC} and LGR\cite{Qin2023GeoTransformerFA}, which rely on sampling correspondence triplets to generate transformation hypotheses, our method proposes hypotheses from single correspondences guided by rotation-consistent features. As shown in Fig.~\ref{fig:2}, our estimator reaches near-optimal performance with only 0.5k hypotheses, while LGR requires around 2k hypotheses to stabilize and RANSAC needs substantially more samples to achieve comparable results. These results demonstrate that leveraging strictly rotation-equivariant features enables more efficient and reliable pose hypothesis generation, leading to improved registration recall with significantly fewer hypotheses.

Overall, our main contributions are summarized as follows:
\begin{itemize}
\item A strictly rotation-equivariant registration framework that explicitly models SO(3) symmetry to resolve the contradiction between pose variation and descriptor invariance.

\item A spherical-harmonics-based backbone that guarantees strict rotation equivariance and produces discriminative rotation-invariant descriptors without relying on approximate invariance.

\item  A transformation estimation strategy enabled by strict $SO(3)$ equivariance, allowing each correspondence to determine a rigid transformation and reducing the RANSAC search space for improved registration accuracy.
\end{itemize}

\section{Related Work}
\label{sec:related}
{\textbf{Patch-wise and Scene-wise Feature Extractors.}}
Early learning-based point cloud descriptors\cite{Yew20183DFeatNetWS,Deng2018PPFNetGC,Deng2018PPFFoldNetUL,Gojcic2018ThePM,Ao2020SpinNetLA} were mainly patch-wise extractors, motivated by traditional handcrafted descriptors\cite{Rusu2009FastPF,Guo2013RoPSAL,Salti2014SHOTUS,Dong2017ANB} and constrained by the limited scalability of early point cloud convolution techniques. These methods focused on designing rotation-invariant local representations, typically through local reference frame (LRF) construction\cite{Gojcic2018ThePM,Ao2020SpinNetLA} or handcrafted invariant encodings such as point pair features (PPFs)\cite{Deng2018PPFNetGC,Deng2018PPFFoldNetUL}. While such strategies remove rotational ambiguity at the input level, LRF estimation is known to be unstable under noise, density variation, and partial observations, and invariant encodings often suppress orientation-dependent geometric cues that are potentially discriminative. With the development of more efficient point cloud backbones\cite{Thomas2019KPConvFA,Choy2019FullyCG}, scene-wise feature extractors\cite{Yu2021CoFiNetRC,Gao2023OAAFormerRA,Qin2023GeoTransformerFA,Chen2024DynamicCT} have become dominant due to their scalability and contextual reasoning capability. These methods often rely on data augmentation to improve robustness to rotations and focus on low-overlap matching via attention mechanisms and coarse-to-fine pipelines. However, augmentation-based training only provides approximate invariance and does not offer theoretical guarantees under arbitrary 3D rotations. To address this limitation, several works\cite{Wang2021YouOH,Wang2023RoRegPP,Yu2023RotationInvariantTF,Yao2024PARENetPR} introduce rotation-invariant or rotation-equivariant architectures based on group convolutions, vector neurons, or invariant convolution operators. Group-based approaches discretize $SO(3)$ and extract features under multiple rotated copies of the input, leading to high computational cost and memory overhead, while lightweight equivariant models often compromise expressive power. In contrast, SHReg models local geometry using irreducible representations of $SO(3)$ via spherical harmonics, ensuring strict rotation equivariance by construction without LRF estimation, rotation discretization, or heavy augmentation, thereby preserving both efficiency and geometric fidelity.

{\textbf{Robust Transformation Estimators.}}
Given noisy correspondences containing many outliers, robust transformation estimation remains a fundamental challenge in point cloud registration. RANSAC\cite{Fischler1981RandomSC} is widely adopted due to its simplicity and general applicability; however, it generates hypotheses by randomly sampling correspondence triplets, which results in a cubic search space and requires a large number of iterations for reliable convergence. Recent learning based estimators\cite{Bai2021PointDSCRP,Lee2021DeepHV,Chen2022SC2PCRAS,Jiang2023RobustOR} improve robustness by modeling spatial consistency among correspondences, enabling more effective inlier identification or correspondence reweighting before applying SVD based transformation estimation. Although these approaches reduce the number of spurious hypotheses, they still fundamentally depend on multiple correspondences to recover rigid transformations. A few works\cite{Xing2024EfficientSC} attempt to estimate rotations from a single correspondence through direct regression, yet such strategies lack explicit geometric constraints and often generalize poorly to unseen rotations. In contrast, SHReg exploits strictly rotation equivariant features governed by irreducible representations of $SO(3)$, allowing each correspondence to directly produce a rigid transformation hypothesis in closed form. By embedding transformation structure into the feature representation itself, SHReg substantially reduces the hypothesis search space while remaining compatible with standard global verification procedures, resulting in improved efficiency and robustness under large rotational variations.

\section{Method}
\label{sec:method}

\subsection{Problem Statement}
Given two partially overlapped point clouds
\[
P=\{p_i\in\mathbb{R}^3\mid i=1,\dots,N\},\qquad
Q=\{q_j\in\mathbb{R}^3\mid j=1,\dots,M\},
\]
our goal is to estimate a rigid transformation
\[
T=(R,t),\qquad R\in SO(3),\; t\in\mathbb{R}^3,
\]
that aligns $P$ to $Q$. 
We propose SHReg, a registration framework based on a strictly $SO(3)$-equivariant spherical-harmonics network $f(\cdot)$, whose output satisfies
\[
f(R\circ X)=\rho(R)\,f(X),\qquad R\in SO(3),
\]
where $\rho(\cdot)$ denotes the $SO(3)$ irrep action, implemented by Wigner-$D$ matrices~\cite{Mattis1981QuantumTO}, on feature representations. Such rotation-equivariant features couple local structure information and orientation information induced by $R$, which can be decoupled into rotation-invariant descriptors for correspondence matching and rotation-equivariant features for transformation estimation.

\begin{figure}[t]
  \centering
  \includegraphics[width=\linewidth, trim=120 0 120 0, clip]{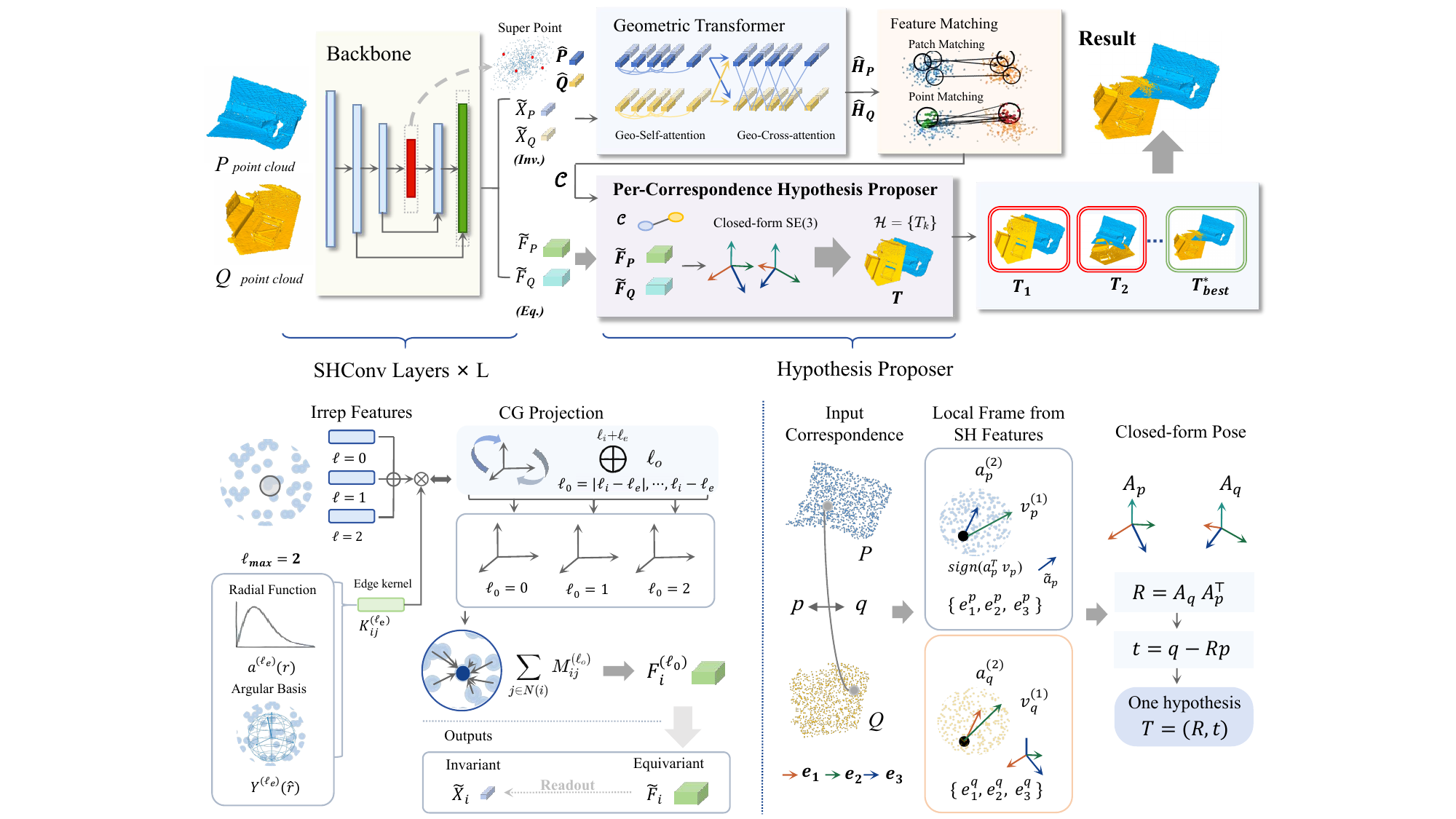}
  \caption{\textbf{Framework of SHReg.} A spherical-harmonics backbone extracts rotation-invariant descriptors and rotation-equivariant features. Correspondences are obtained in a coarse-to-fine manner. Each correspondence generates one $SE(3)$ hypothesis, and the best hypothesis is selected as the final transformation.}
  \label{fig:3}
\end{figure}

The overall framework is illustrated in Fig.~\ref{fig:3}. Given $P$ and $Q$, a hierarchical spherical-harmonics backbone extracts both patch-level and point-level features as $SO(3)$ irreps, and derives rotation-invariant descriptors for coarse-to-fine correspondence estimation. Then, the feature-based hypothesis proposer leverages the matched rotation-equivariant features $\tilde{F}_P$ and $\tilde{F}_Q$ to generate multiple reliable $SE(3)$ hypotheses from individual correspondences in closed form. The best hypothesis is selected through geometric verification and refined to produce the final alignment $(R^\ast,t^\ast)$.

\subsection{Strictly $SO(3)$-Equivariant Feature Backbone}
\label{subsec:backbone}

\subsubsection{Spherical-Harmonic Irreps Representation}
A core component of SHReg is to encode point-wise features as a direct sum of irreducible representations (irreps) of $SO(3)$.
For each point $p_i$, we decompose its feature into type-$\ell$ blocks:
\begin{equation}
F_i=\bigoplus_{\ell=0}^{\ell_{\max}}F_i^{(\ell)},\qquad
F_i^{(\ell)}\in\mathbb{R}^{C_\ell\times(2\ell+1)}.
\label{eq:irrep_decomp}
\end{equation}
Here, $\oplus$ denotes the direct sum of $SO(3)$ irrep blocks rather than ordinary summation; each block contains $C_\ell$ channels with representation dimension $2\ell+1$.
The order $\ell=0$ corresponds to scalar channels and $\ell>0$ corresponds to higher-order geometric tensors.
Under any rotation $R\in SO(3)$ applied to the input point cloud, each block transforms equivariantly:
\begin{equation}
F_i^{(\ell)}(R\circ P)=F_i^{(\ell)}(P)\,D^{(\ell)}(R)^\top,
\label{eq:wignerD_equiv}
\end{equation}
where $D^{(\ell)}(R)$ is the Wigner-$D$ matrix of order $\ell$.
This irrep structure provides strict $SO(3)$ equivariance without pose normalization. We provide a rigorous proof in the supplementary material that our spherical-harmonics encoding, tensor-product with Clebsch--Gordan projection, linear neighborhood aggregation, and scalar-gated nonlinearity jointly preserve $SO(3)$-equivariance across layers.

\subsubsection{$SO(3)$-Equivariant Spherical-Harmonic Convolution}
Given a point cloud $P$, we construct a neighborhood $\mathcal{N}_i$ for each point $p_i$.
For an edge $(i,j)$, we define the relative displacement
\begin{equation}
r_{ij}=p_j-p_i,\quad
r=\|r_{ij}\|,\quad
\hat r_{ij}=\frac{r_{ij}}{r}.
\label{eq:relative_geom}
\end{equation}
The unit direction $\hat r_{ij}$ is lifted to an equivariant angular basis by spherical harmonics $Y^{(\ell)}(\hat r_{ij})$,
while the distance $r$ is used to generate learnable radial weights:
\begin{equation}
a^{(\ell)}(r)=\mathrm{MLP}_\ell(\phi(r)),
\label{eq:radial_filter}
\end{equation}
where $\phi(\cdot)$ denotes a radial basis function (RBF) expansion.
The edge kernel is therefore given by $K_{ij}^{(\ell_e)}=a^{(\ell_e)}(r)Y^{(\ell_e)}(\hat r_{ij})$, where the radial MLP only takes the scalar distance $r$, while $Y^{(\ell_e)}$ provides the angular equivariant basis.

We define message passing via a tensor-product convolution followed by a Clebsch--Gordan (CG) projection:
\begin{equation}
M_{ij}^{(\ell_o)}
=\sum_{\ell_i,\ell_e}
C_{\ell_i,\ell_e\rightarrow\ell_o}
\Big(
F_j^{(\ell_i)}
\otimes
\big(a^{(\ell_e)}(r_{ij})\,Y^{(\ell_e)}(\hat r_{ij})\big)
\Big),
\label{eq:tp_conv}
\end{equation}
where $C_{\ell_i,\ell_e\rightarrow\ell_o}$ is the fixed CG projection from type-$\ell_i$ and type-$\ell_e$ irreps to a valid type-$\ell_o$ output irrep, with $|\ell_i-\ell_e|\leq \ell_o\leq \ell_i+\ell_e$.
Only radial functions and channel mixing weights are learnable.
Aggregating messages over the neighborhood yields the updated feature:
\begin{equation}
F_i^{(\ell_o)\prime}=\sum_{p_j\in\mathcal{N}_i}M_{ij}^{(\ell_o)}.
\label{eq:agg_update}
\end{equation}
We use invariant $\ell=0$ gates to modulate higher-order channels, so the scalar-gated nonlinearity preserves the corresponding $SO(3)$ transformation rules.
Stacking such layers forms a hierarchical backbone that outputs multi-resolution rotation-equivariant features $\tilde F$,
from which rotation-invariant descriptors $\tilde X$ are further derived for matching.

\subsubsection{Rotation-Invariant Descriptor Readout}
For correspondence matching, we remove the representation dimension within each irrep block using an invariant norm:
\begin{equation}
X_{i,c}^{(\ell)}=\sum_{m=-\ell}^{\ell}\big|F_{i,c}^{(\ell,m)}\big|^2,
\label{eq:invariant_norm}
\end{equation}
where $c=1,\dots,C_\ell$ denotes the channel index.
Concatenating invariants across all orders produces the final descriptor:
\begin{equation}
d_i=\mathrm{Concat}\big(\{X_i^{(\ell)}\}_{\ell=0}^{\ell_{\max}}\big).
\label{eq:final_descriptor}
\end{equation}
The descriptor dimension before readout layers is therefore $\sum_{\ell=0}^{\ell_{\max}}C_\ell$.
Since $D^{(\ell)}(R)$ is orthogonal, the $\ell_2$ norm in Eq.~\eqref{eq:invariant_norm} is preserved under arbitrary rotations,
and thus $d_i$ is strictly rotation-invariant. A derivation of the strict invariance of Eqs.~\eqref{eq:invariant_norm}--\eqref{eq:final_descriptor} is provided in the supplementary.

\subsection{Coarse-to-Fine Correspondence Matching}
\label{subsec:matching}

Since the backbone produces both patch-level and point-level features, we adopt a coarse-to-fine matching strategy to suppress non-overlapping regions and establish accurate correspondences.

\subsubsection{Superpoint Matching}

We first downsample the input point clouds into sparse superpoints $\hat P$ and $\hat Q$.
Their rotation-invariant features $\hat X_P$ and $\hat X_Q$ are fed into a Geometric Transformer module, which captures global context via iterative self- and cross-attention.
The refined superpoint features $\hat H_P$ and $\hat H_Q$ are then used to compute a similarity matrix $S$.
The top-$k$ reliable superpoint correspondences are selected as
\begin{equation}
\hat C=\{(\hat p_{x_i},\hat q_{y_i})\}.
\label{eq:super_corr}
\end{equation}

\subsubsection{Point Matching}

For each matched superpoint pair, we further search point-level correspondences within their associated groups $G^P_{x_i}$ and $G^Q_{y_i}$.
Following~\cite{Lindenberger2023LightGlueLF}, we disentangle feature similarity and saliency by introducing a matchability head $W_m$ and a saliency head $W_s$.

The similarity matrix is computed as
\begin{equation}
M_{x_i,y_i}=(W_m X^P_{x_i})^\top (W_m X^Q_{y_i})/\sqrt{D},
\label{eq:point_sim}
\end{equation}
where $D$ denotes the feature dimension.
The saliency scores are predicted by
\begin{equation}
\sigma^P_{x_i}=\mathrm{Sigmoid}(W_s X^P_{x_i}),\qquad
\sigma^Q_{y_i}=\mathrm{Sigmoid}(W_s X^Q_{y_i}).
\label{eq:saliency}
\end{equation}
We obtain a soft assignment matrix
\begin{equation}
Z_i=\sigma^P_{x_i}\sigma^Q_{y_i}\,\varphi(M_{x_i,y_i}),
\label{eq:soft_assign}
\end{equation}
where $\varphi$ denotes softmax normalization.
Finally, the top-ranked point correspondences across all matched superpoint pairs are selected:
\begin{equation}
C=\{(\tilde p_{x_j},\tilde q_{y_j})\}.
\label{eq:point_corr}
\end{equation}

\subsection{Feature-Based Hypothesis Proposer}

Given a single correspondence $(p\leftrightarrow q)$, we generate one $SE(3)$ hypothesis $T=(R,t)$ in closed form from the matched rotation-equivariant features.
The key idea is to construct a local orthonormal frame at each point using different irreps: $\ell=2$ captures a stable principal axis, while $\ell=1$ resolves its orientation.

\subsubsection{Principal axis from $\ell=2$.}
We first aggregate $\ell=2$ features across channels:
\begin{equation}
u_p=w_2^\top\mathbf F_p^{(2)},\qquad
u_q=w_2^\top\mathbf F_q^{(2)}, \quad u_p,u_q\in\mathbb{R}^5.
\end{equation}
Each vector is mapped to a symmetric trace-free tensor:
\begin{equation}
S_p=\Phi(u_p),\qquad
S_q=\Phi(u_q).
\end{equation}
The dominant eigenvectors of $S_p$ and $S_q$, denoted by $a_p$ and $a_q$, define unsigned principal axes.

\subsubsection{Axis orientation from $\ell=1$.}
To resolve the sign ambiguity, we extract a direction from $\ell=1$ features:
\begin{equation}
v_p=w_1^\top\mathbf F_p^{(1)},\qquad
v_q=w_1^\top\mathbf F_q^{(1)}.
\end{equation}
The axis orientation is determined by
\begin{equation}
\tilde a_p=\mathrm{sign}(a_p^\top v_p)\,a_p,\qquad
\tilde a_q=\mathrm{sign}(a_q^\top v_q)\,a_q.
\end{equation}

\subsubsection{Local frame construction.}
We construct right-handed local frames:
\begin{equation}
e_1=\mathrm{normalize}(\tilde a),\quad
e_2=\mathrm{normalize}\big(v-(e_1^\top v)e_1\big),\quad
e_3=e_1\times e_2,
\end{equation}
forming orthonormal matrices $A_p,A_q\in SO(3)$.

\subsubsection{Closed-form transformation.}
The rotation aligning the two frames is
\begin{equation}
R=A_qA_p^\top,
\end{equation}
and the translation is computed as
\begin{equation}
t=q-Rp.
\end{equation}
Therefore, each correspondence directly produces one $SE(3)$ hypothesis without multi-point minimal solvers or SVD~\cite{Kabsch1976ASF}. We detail the frame construction from $\ell=1,2$ features and degenerate-case checks in the supplementary.

\subsection{Hypothesis Verification and Selection}

We adopt a local-to-global hypothesis verification scheme. Given the correspondence set $\mathcal C$, the local phase generates a hypothesis set $\mathcal H=\{T_k=(R_k,t_k)\}$ by applying the above closed-form solution to each correspondence. Since each hypothesis is inferred from matched rotation-equivariant features encoding local structural orientation, the resulting transformations are already locally consistent.

In the global phase, we select the transformation that maximizes inlier support over all correspondences:
\begin{equation}
T^*=\arg\max_{T_k\in\mathcal H}
\sum_{(\tilde p_{x_j},\tilde q_{y_j})\in\mathcal C}
\llbracket
\|R_k \tilde p_{x_j}+t_k-\tilde q_{y_j}\|_2^2<\tau
\rrbracket,
\end{equation}
where $\llbracket\cdot\rrbracket$ denotes the Iverson bracket and $\tau$ is the inlier threshold. Compared with sampling-based estimators such as RANSAC~\cite{Fischler1981RandomSC}, SHReg generates one closed-form hypothesis per correspondence, substantially reducing the hypothesis search space while preserving high registration accuracy.

\subsection{Loss Function}
\label{subsec:loss}

We follow PARE-Net~\cite{Yao2024PARENetPR} and train SHReg with three losses,
$\mathcal{L}=\mathcal{L}_c+\mathcal{L}_f+\mathcal{L}_r$,
including a superpoint matching loss $\mathcal{L}_c$, a point matching loss $\mathcal{L}_f$, and a contrastive rotation loss $\mathcal{L}_r$.
Specifically, $\mathcal{L}_c$ supervises rotation-invariant superpoint descriptors using the overlap-aware circle loss, $\mathcal{L}_f$ optimizes fine-level correspondences by supervising the soft assignment matrix and saliency prediction within ground-truth superpoint pairs, and $\mathcal{L}_r$ regularizes rotation-equivariant features to be consistent under the ground-truth rotation via a margin-based contrastive objective.
For completeness, we refer readers to PARE-Net~\cite{Yao2024PARENetPR} for the detailed formulations and training settings of these loss terms.

\section{Experiments}
To evaluate the effectiveness of SHReg, we compare it with state-of-the-art point cloud registration methods on 3DMatch~\cite{Zeng20163DMatchLL}, 3DLoMatch~\cite{Huang2020PREDATORRO}, and KITTI Odometry~\cite{Geiger2012AreWR}. We also conduct ablation studies to analyze the contribution of each component. All experiments are conducted on a workstation with an RTX 4090 GPU and an Intel Xeon(R) Platinum 8352V CPU. More implementation details are provided in the Appendix.

\subsection{Indoor Dataset: 3DMatch \& 3DLoMatch}

{\textbf{Dataset.}}
3DMatch~\cite{Zeng20163DMatchLL} and 3DLoMatch~\cite{Huang2020PREDATORRO} are indoor RGB-D benchmarks. 3DMatch contains 62 scenes, with 46, 8, and 8 scenes used for training, validation, and testing, respectively. It consists of point cloud pairs with overlap ratios greater than 30\%. 3DLoMatch is used only for testing and contains more challenging pairs with overlap ratios between 10\% and 30\%. We follow the standard protocols~\cite{Huang2020PREDATORRO,Yao2024PARENetPR}. Unless otherwise specified, 5000 keypoints are sampled per fragment to establish correspondences.

{\textbf{Baselines.}}
We compare SHReg with strong baselines from three categories. The first category includes rotation-sensitive scene-level matchers and descriptors, such as FCGF~\cite{Choy2019FullyCG}, Predator~\cite{Huang2020PREDATORRO}, CoFiNet~\cite{Yu2021CoFiNetRC}, GeoTransformer~\cite{Qin2023GeoTransformerFA}, and PEAL~\cite{Yu2023PEALPE}. The second category contains rotation-robust scene-level methods, including YOHO~\cite{Wang2021YouOH}, RoReg~\cite{Wang2023RoRegPP}, RoITR~\cite{Yu2023RotationInvariantTF}, and PARE-Net~\cite{Yao2024PARENetPR}. The third category consists of patch-level rotation-invariant descriptors, including SpinNet~\cite{Ao2020SpinNetLA} and BUFFER~\cite{Ao2023BUFFERBA}. For methods that require a robust estimator, we use the same Open3D RANSAC implementation with optimized hyperparameters for fair comparison.

{\textbf{Metrics.}}
Following~\cite{Huang2020PREDATORRO,Qin2023GeoTransformerFA}, we evaluate registration performance using Inlier Ratio (IR), Feature Matching Recall (FMR), and Registration Recall (RR). A correspondence is considered correct if its residual distance under the ground-truth transformation is smaller than $\tau_c$. IR denotes the ratio of correct correspondences, FMR is the percentage of scan pairs whose IR exceeds 5\%, and RR measures the fraction of successfully registered pairs under threshold $\tau_r$. For rotated datasets, we additionally report Rotation Error (RE), Translation Error (TE), and Transformation Recall (TR) following RoReg~\cite{Wang2023RoRegPP}. Detailed definitions of these metrics are provided in the Appendix.

\begin{table*}[t]
\centering
\caption{Evaluation results on 3DMatch and 3DLoMatch.
Methods with the RANSAC estimator are marked by $\diamond$,
which exploit 5000 points to establish correspondences.}
\vspace{-8pt}
\label{tab:1}
\resizebox{\textwidth}{!}{
\begin{tabular}{l c cccc cccc}
\toprule
\multirow{2}{*}{Method} & 
\multirow{2}{*}{Size (MB)} &
\multicolumn{4}{c}{3DMatch} &
\multicolumn{4}{c}{3DLoMatch} \\
\cmidrule(lr){3-6} \cmidrule(lr){7-10}
& &
FMR (\%$\uparrow$) & IR (\%$\uparrow$) & RR (\%$\uparrow$) & Time (s$\downarrow$) &
FMR (\%$\uparrow$) & IR (\%$\uparrow$) & RR (\%$\uparrow$) & Time (s$\downarrow$) \\
\midrule

FCGF$\diamond$~\cite{Choy2019FullyCG}      
& 8.76 
& 94.7 & 31.1 & 82.8 & \textbf{0.12}
& 59.4 & 9.8 & 38.0 & \textbf{0.13} \\

SpinNet$\diamond$~\cite{Ao2020SpinNetLA} 
& \underline{1.41} 
& 97.6 & 47.5 & 88.6 & 9.85 
& 75.3 & 20.5 & 59.8 & 9.03 \\

YOHO~\cite{Wang2021YouOH}                
& 12.38 
& 98.2 & 64.4 & 90.8 & 2.81 
& 78.9 & 25.9 & 66.0 & 2.62 \\

Predator$\diamond$~\cite{Huang2020PREDATORRO} 
& 7.43 
& 96.6 & 58.0 & 89.0 & 0.64 
& 78.2 & 26.7 & 64.4 & 0.47 \\

GeoTrans~\cite{Qin2023GeoTransformerFA}        
& 9.83 
& 98.1 & 70.9 & 92.4 & 0.18 
& 87.4 & 43.5 & 74.3 & 0.17 \\

RoReg~\cite{Wang2023RoRegPP}              
& 12.71 
& 98.2 & \underline{81.6} & 93.0 & 2.27 
& 82.3 & 39.6 & 70.1 & 2.10 \\

BUFFER~\cite{Ao2023BUFFERBA}            
& \textbf{0.92} 
& -- & -- & 92.1 & 0.22 
& -- & -- & 70.3 & 0.21 \\

RoITr$\diamond$~\cite{Yu2023RotationInvariantTF}    
& 10.10 
& 98.0 & \textbf{82.4} & 91.9 & 0.36 
& \textbf{89.2} & \textbf{54.6} & 74.1 & 0.34 \\

PEAL\cite{Yu2023PEALPE}                
& 9.83 
& 98.4 & 71.0 & 94.2 & 1.46 
& 88.3 & 46.0 & 78.8 & 1.19 \\

PARE-Net\cite{Yao2024PARENetPR}               
& 3.84 
& \underline{98.5} & 76.9 & \underline{95.0} & \underline{0.17}
& 88.3 & 47.5 & \underline{80.5} & \underline{0.17} \\

\midrule

\textbf{Ours} 
& 9.52 
& \textbf{98.5} & 78.6 & \textbf{95.4} & 0.28
& \underline{88.6} & \underline{48.8} & \textbf{82.4} & 0.28 \\

\bottomrule
\end{tabular}
}
\vspace{-15pt}
\end{table*}

\vspace{-10pt}
\begin{table*}[h]
\centering
\caption{Evaluation results on 3DLoMatch and Rotated 3DLoMatch. 
For easy comparison of the impact of rotation, the numerical changes w.r.t. TR on the rotated dataset are annotated in the top right corner of each value.
Methods with rotation invariant/equivariant designs are marked by * symbol. }
\vspace{-8pt}
\label{tab:2}
\setlength{\tabcolsep}{5pt}
\footnotesize
\resizebox{\textwidth}{!}{
\begin{tabular}{l c ccc ccc}
\toprule
\multirow{2}{*}{Method} &
\multirow{2}{*}{Size (MB)} &
\multicolumn{3}{c}{3DLoMatch} &
\multicolumn{3}{c}{Rotated 3DLoMatch} \\
\cmidrule(lr){3-5} \cmidrule(lr){6-8}
& &
RE ($^\circ\downarrow$) & TE (cm$\downarrow$) & TR (\%$\uparrow$) &
RE ($^\circ\downarrow$) & TE (cm$\downarrow$) & TR (\%$\uparrow$) \\
\midrule
FCGF~\cite{Choy2019FullyCG}         & 8.76  & 4.84 & 12.87 & 39.6 &
4.74 & 13.39 & $24.5^{-15.1}$ \\

Predator~\cite{Huang2020PREDATORRO} & 7.43  & 3.61 & 10.65 & 65.6 &
3.55 & 10.30 & $64.0^{-1.6}$ \\

GeoTrans~\cite{Qin2023GeoTransformerFA} & 9.83  & 2.91 & 8.71  & 75.4 &
2.94 & 8.85  & $72.6^{-2.8}$ \\

PEAL~\cite{Yu2023PEALPE}         
& 9.83  
& \textbf{2.84} & \textbf{8.64} & 81.2 &
\underline{2.86} & \textbf{8.53} & $78.7^{-2.5}$ \\

\midrule

YOHO*~\cite{Wang2021YouOH}        
& 12.38 & 3.54 & 10.34 & 66.6 &
3.61 & 10.16 & $67.1^{+0.5}$ \\

RoReg*~\cite{Wang2023RoRegPP}      
& 12.71 & 3.01 & 9.26 & 71.3 &
3.03 & 9.28 & $71.0^{-0.3}$ \\

BUFFER*~\cite{Ao2023BUFFERBA}    
& \textbf{0.92}  & 3.03 & 9.86 & 74.4 &
3.02 & 9.99 & $74.7^{+0.3}$ \\

RoITr*~\cite{Yu2023RotationInvariantTF}      
& 10.10 & 2.95 & 9.03 & 75.1 &
2.97 & 9.08 & $75.5^{+0.4}$ \\

PARE-Net*~\cite{Yao2024PARENetPR}            
& \underline{3.84} 
& \underline{2.87} & \underline{8.83} & \underline{81.3} &
\textbf{2.84} & \underline{8.71} & $\underline{81.8^{+0.5}}$ \\

\textbf{Ours*}           
& 9.52  
& 2.92 & 8.96 & \textbf{83.0} &
2.88 & 8.94 & $\mathbf{83.3^{+0.3}}$ \\

\bottomrule
\end{tabular}
}
\end{table*}
\vspace{-9pt}

{\textbf{Quantitative results.}}
We report the results on 3DMatch and 3DLoMatch in Table~\ref{tab:1}. SHReg achieves consistently strong performance on both datasets, especially on 3DLoMatch where low overlap makes correspondence quality and hypothesis generation more challenging. Compared with rotation-sensitive methods such as Predator\cite{Huang2020PREDATORRO} and GeoTransformer\cite{Qin2023GeoTransformerFA}, SHReg exhibits better robustness under hard cases since it does not rely on extensive rotation augmentation to approximate invariance. Compared with rotation-invariant/equivariant competitors such as YOHO\cite{Wang2021YouOH}, RoReg\cite{Wang2023RoRegPP}, RoITr\cite{Yu2023RotationInvariantTF}, and PARE-Net\cite{Yao2024PARENetPR}, SHReg yields improved IR and RR while maintaining high efficiency. This advantage mainly comes from two aspects: (1) strictly $SO(3)$-equivariant feature learning provides stable local structure encoding under arbitrary rotations, and (2) equivariant features allow transformation hypotheses to be proposed with significantly reduced hypothesis complexity compared with triplet-based sampling.

{\textbf{Experiments on the Rotated Dataset.}}
To test robustness against unseen rotations, we follow prior works~\cite{Wang2023RoRegPP,Yu2023RotationInvariantTF,Yao2024PARENetPR} to evaluate all methods on the rotated 3DLoMatch benchmark, where arbitrary full-range rotations are independently applied to each fragment of a pair.
This setting explicitly tests whether a method is intrinsically rotation-invariant/equivariant rather than relying on limited coverage of rotation augmentation. 
The results are shown in Table~\ref{tab:2}.
Rotation-sensitive methods such as FCGF\cite{Choy2019FullyCG}, Predator\cite{Huang2020PREDATORRO}, GeoTransformer\cite{Qin2023GeoTransformerFA},
and PEAL\cite{Yu2023PEALPE} present noticeable performance drops on the rotated benchmark, indicating that approximate invariance obtained by augmentation generalizes poorly to unseen rotations. In contrast, SHReg maintains stable performance and even shows slight improvements in TR, demonstrating that strict equivariance/invariance by design provides reliable robustness under severe pose variations. Moreover, SHReg surpasses previous rotation-robust methods such as YOHO\cite{Wang2021YouOH}, RoReg\cite{Wang2023RoRegPP}, RoITr\cite{Yu2023RotationInvariantTF} and PARE-Net\cite{Yao2024PARENetPR}, suggesting that spherical-harmonics-based irreps better preserve discriminative structural information.

\begin{figure}[t]
\vspace{-6pt}
  \centering
  \includegraphics[width=\linewidth]{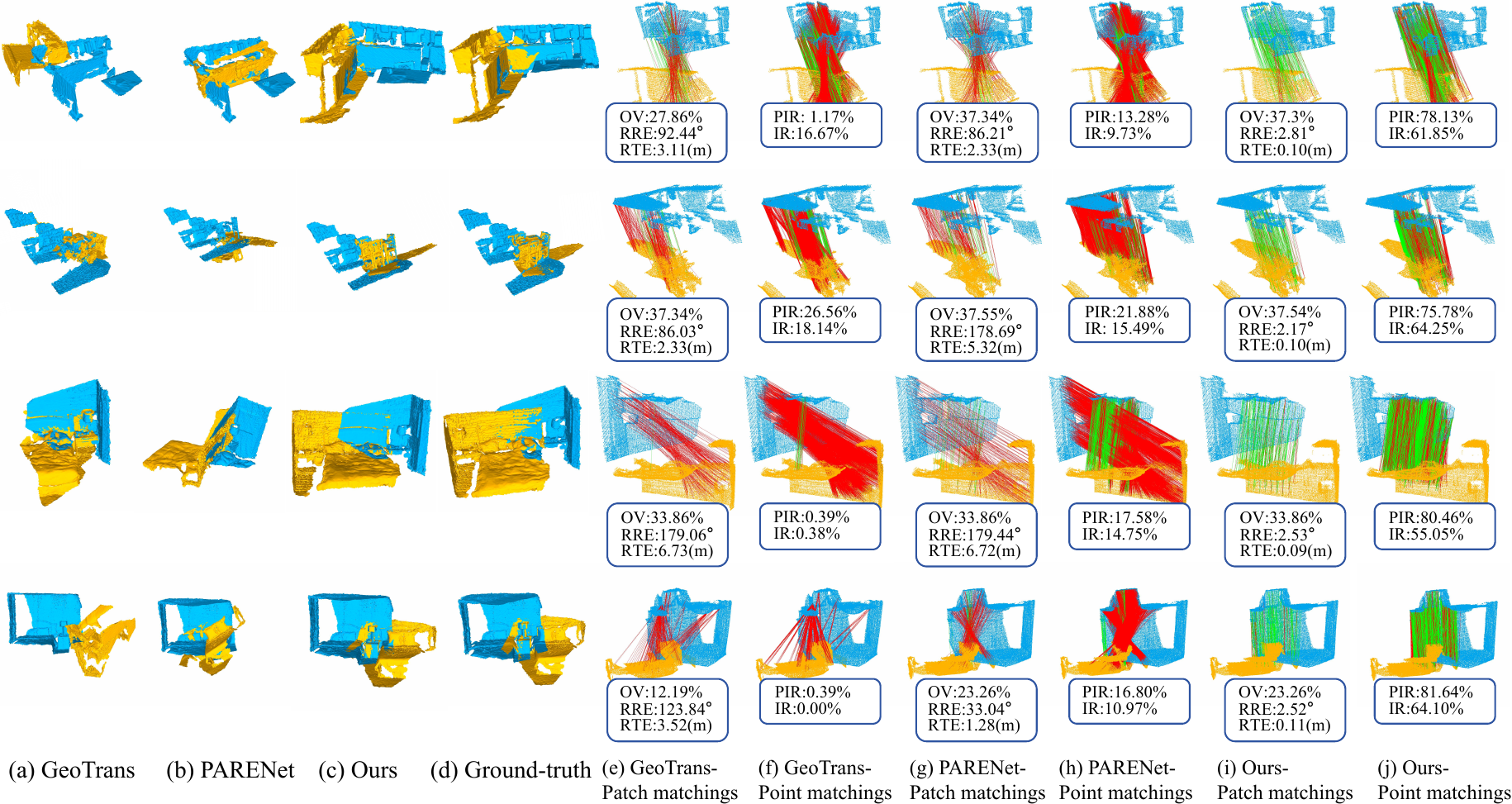}
  \caption{Qualitative results. Our method can successfully align low-overlapped pairs with higher IR and PIR (Patch IR). Moreover, GeoTrans\cite{Qin2023GeoTransformerFA} and PARE-Net\cite{Yao2024PARENetPR} incorrectly align point clouds with opposite orientations (the first and second rows), our method successfully addresses this issue because the equivariant features contain orientation information of local structure, alleviating this ambiguity.}
  \label{fig:5}
\vspace{-10pt}
\end{figure}

\textbf{Qualitative results.}
We present representative registration results in Fig.~\ref{fig:5}. 
SHReg produces more accurate patch-level and point-level correspondences and successfully aligns challenging fragment pairs with low overlap ratios. 
We observe that rotation-sensitive methods such as GeoTransformer~\cite{Qin2023GeoTransformerFA} may incorrectly align scans with opposite orientations under severe pose variations. 
This ambiguity typically arises when point-to-point constraints alone are insufficient to disambiguate local structural symmetries, especially under large rotations.
In contrast, SHReg successfully resolves such flipped alignments. 
This is because the learned rotation-equivariant features explicitly encode the orientation of local geometric structures, allowing our hypothesis proposer to construct consistent local frames from matched correspondences and generate geometrically valid transformations. 
Additional qualitative results and failure case analysis are provided in the Appendix.

\begin{table}[t]

\centering
\caption{Evaluation results on KITTI Odometry.}
\vspace{-8pt}
\label{tab:3}
\setlength{\tabcolsep}{17pt}
\footnotesize
\resizebox{\textwidth}{!}{
\begin{tabular}{l c c c c c}
\toprule
Method & Size(MB) & RE ($^\circ$) & TE (cm) & TR (\%) & Time (s) \\
\midrule
FCGF\cite{Choy2019FullyCG}          & 8.76  & 0.30 & 9.5 & 96.6 & -- \\
D3Feat\cite{Bai2020D3FeatJL}      & 14.08 & 0.30 & 7.2 & \textbf{99.8} & -- \\
Predator\cite{Huang2020PREDATORRO}  & 22.77 & 0.27 & 6.8 & \textbf{99.8} & 0.77 \\
SpinNet\cite{Ao2020SpinNetLA}    & \underline{1.41}  & 0.47 & 9.9 & 99.1 & 16.24 \\
CoFiNet\cite{Yu2021CoFiNetRC}    & 5.48  & 0.41 & 8.2 & \textbf{99.8} & 0.59 \\
GeoTrans\cite{Qin2023GeoTransformerFA}  & 25.50 & \textbf{0.23} & 6.2 & \textbf{99.8} & 0.26 \\
BUFFER\cite{Ao2023BUFFERBA}      & \textbf{0.92} & 0.26 & 7.1 & \textbf{99.8} & 0.27 \\
PARE-Net\cite{Yao2024PARENetPR}            & 2.08  & \textbf{0.23} & \underline{4.9} & \textbf{99.8} & \textbf{0.21} \\
\textbf{Ours}             &24.82   &\textbf{0.23}  &\textbf{4.7}  &\textbf{99.8}  &\underline{0.24}  \\

\bottomrule
\end{tabular}}

\end{table}

\subsection{Outdoor Dataset: KITTI Odometry}

{\textbf{Dataset.}}
We further evaluate SHReg on the KITTI Odometry dataset~\cite{Geiger2012AreWR}, which contains outdoor LiDAR scans captured in urban driving environments. Following previous works~\cite{Choy2019FullyCG, Qin2023GeoTransformerFA, Yao2024PARENetPR}, we use sequences 0--5 for training, 6--7 for validation, and 8--10 for testing.
SHReg is trained on indoor 3DMatch and directly evaluated on KITTI without fine-tuning, demonstrating its cross-domain generalization ability.

{\textbf{Metrics.}}
We adopt the standard metrics Rotation Error (RE), Translation Error (TE), and Transformation Recall (TR).
Following~\cite{Choy2019FullyCG, Qin2023GeoTransformerFA, Yao2024PARENetPR}, TR is computed with a rotation threshold of $5^\circ$ and a translation threshold of 2m.

{\textbf{Quantitative results.}}
Quantitative results on KITTI are reported in
Table~\ref{tab:3}.
On this benchmark, many methods have reached performance saturation, while SHReg
achieves highly competitive accuracy with improved efficiency.
These results validate that the proposed strictly equivariant representation can
generalize well from indoor RGB-D scans to outdoor LiDAR scenarios.

\subsection{Ablation Study}

To better understand the contribution of each component in SHReg, we conduct ablation experiments on the 3DMatch and 3DLoMatch benchmarks. The quantitative results are summarized in Table~\ref{tab:4}.

{\textbf{Ablation of backbone.}}
We first evaluate a rotation-sensitive baseline backbone that does not explicitly enforce rotational equivariance. We then introduce the key components of the proposed spherical-harmonics-based equivariant backbone, including spherical harmonic (SH) directional encoding, Clebsch--Gordan (CG) projection, gated nonlinearity, and rotation-invariant descriptor readout. The results show that SH representations consistently improve feature quality. In particular, removing the invariant readout significantly degrades performance, since equivariant features still contain orientation-dependent components and are not directly suitable for correspondence matching. Removing CG projection or the gating mechanism also leads to noticeable drops, indicating the importance of valid irrep coupling and stable equivariant nonlinear transformations. The full SHReg backbone achieves the best performance on both datasets, demonstrating the effectiveness of strictly $SO(3)$-equivariant feature learning.

{\textbf{Ablation of pose estimator.}}
We further evaluate different pose estimation strategies using the same correspondence sets, including standard RANSAC with randomly sampled correspondence triplets, a patch-based estimator similar to LGR, and our single-correspondence hypothesis proposer. Since all estimators share identical correspondences, FMR and IR remain unchanged, while RR reflects the pose estimation quality. As shown in Table~\ref{tab:4}, our hypothesis proposer achieves the highest RR on both datasets, especially on 3DLoMatch. This is because low-overlap pairs make triplet-based sampling more likely to produce unstable hypotheses, whereas SHReg constructs transformations directly from individual correspondences using orientation information encoded in rotation-equivariant features. This reduces the reliance on extensive random sampling and improves pose estimation robustness.

\begin{table*}[t]

\centering
\caption{Ablation experiments of SHReg.}
\vspace{-8pt}
\label{tab:4}
\setlength{\tabcolsep}{7pt}
\footnotesize
\resizebox{\textwidth}{!}{
\begin{tabular}{l l ccc ccc}
\toprule
\multirow{2}{*}{Component} & \multirow{2}{*}{Method} &
\multicolumn{3}{c}{3DMatch} & \multicolumn{3}{c}{3DLoMatch} \\
\cmidrule(lr){3-5} \cmidrule(lr){6-8}
& &
FMR (\%$\uparrow$) & IR (\%$\uparrow$) & RR (\%$\uparrow$) &
FMR (\%$\uparrow$) & IR (\%$\uparrow$) & RR (\%$\uparrow$) \\
\midrule

\multirow{5}{*}{Backbone}
& Rotation-sensitive baseline (no SH/irreps)
& 98.0 & 71.0 & 93.6 & 86.7 & 41.5 & 78.2 \\

& SH encoding w/o CG projection
& 98.2 & 73.5 & 94.2 & 87.5 & 43.5 & 79.3 \\

& SH + CG, w/o Gate nonlinearity
& \underline{98.3} & \underline{75.8} & \underline{94.7} & \underline{88.0} & \underline{45.6} & \underline{80.1} \\

& SH + CG + Gate, w/o invariant readout
& 97.6 & 68.0 & 92.8 & 84.0 & 36.5 & 75.5 \\

& \textbf{Full SHReg backbone}
& \textbf{98.5} & \textbf{78.6} & \textbf{95.4} & \textbf{88.6} & \textbf{48.8} & \textbf{82.4} \\

\midrule

\multirow{3}{*}{Pose Estimator}
& RANSAC (triplet sampling)
& \textbf{98.5} & \textbf{78.6} & 94.0 & \textbf{88.6} & \textbf{48.8} & 79.5 \\

& LGR / patch-based estimator
& \textbf{98.5} & \textbf{78.6} & \underline{94.6} & \textbf{88.6} & \textbf{48.8} & \underline{81.0} \\

& \textbf{Ours} (single-correspondence proposer)
& \textbf{98.5} & \textbf{78.6} & \textbf{95.4} & \textbf{88.6} & \textbf{48.8} & \textbf{82.4} \\

\bottomrule
\end{tabular}}
\end{table*}

\section{Conclusion}

In this paper, we presented SHReg, a point cloud registration framework built upon strictly $SO(3)$-equivariant spherical-harmonic representations. By modeling local features as irreducible representations of $SO(3)$, SHReg guarantees consistent feature transformations under arbitrary rotations without relying on local reference frames or rotation augmentation. The proposed backbone produces both rotation-invariant descriptors for reliable correspondence matching and rotation-equivariant features that preserve structural orientation cues. Leveraging this property, we introduced a correspondence-driven hypothesis proposer that estimates rigid transformations directly from individual correspondences, significantly reducing the hypothesis search space compared with conventional sampling-based estimators. Extensive experiments on indoor and outdoor benchmarks demonstrate that SHReg achieves strong registration accuracy and robustness, particularly under large rotational variations.


\section*{Acknowledgments}
This work was partially supported by the National Natural Science Foundation of China under Grant No.~62373164, the Project of Central Government Guides Local Program under Grant No.~YDZX2024075, the Natural Science Foundation of Shandong Province Youth Program under Grant No.~ZR2025QC2246Z, the Taishan Scholars Program of Shandong Province under Grant No.~tsqn202507271, and the Taishan Experts Program under Grant No.~tscy20241154.

%
\bibliographystyle{splncs04}
\bibliography{main}
\end{document}